%% file: main.tex
\documentclass[letterpaper,twocolumn,10pt]{article}
\usepackage{usenix2019_v3}

\usepackage{amsmath}
\usepackage{booktabs}
\usepackage{hyperref}
\usepackage{caption}
\usepackage{subcaption}
\newcommand{\bnns}{RA-BNN\xspace}
\newcommand{\grows}{Early Growth\xspace}
\usepackage{times}

\usepackage{graphicx}

\usepackage{times}
\usepackage{epsfig}
\usepackage{graphicx}
\usepackage{amsmath}
\usepackage{amssymb}
\usepackage{multirow}
\usepackage{xspace}
\usepackage{dblfloatfix}

\usepackage{bm}
\usepackage{cleveref}
\usepackage{authblk}
\usepackage{url}

\input{math_commands.tex}

\title{RA-BNN: Constructing \underline{R}obust \& \underline{A}ccurate \underline{B}inary \underline{N}eural \underline{N}etwork to Simultaneously Defend Adversarial Bit-Flip Attack and Improve Accuracy}

\author[1]{Adnan Siraj Rakin}
\author[1]{Li Yang} 
\author[1]{Jingtao Li} 
\author[2]{ Fan Yao}
\author[1]{Chaitali Chakrabarti}
\author[1]{Yu Cao}
\author[1]{Jae-sun Seo}
\author[1]{Deliang Fan}

\affil[1]{Department of Electrical,Computer and Energy Engineering, Arizona State University}
\affil[2]{Department of Electrical and Computer Engineering, University of Central Florida}

\begin{document}

\date{}


\maketitle

\begin{abstract}
Recently developed adversarial weight attack, a.k.a. bit-flip attack (BFA), has shown enormous success in compromising Deep Neural Network (DNN) performance with an extremely small amount of model parameter perturbation. To defend against this threat, we propose \textbf{\bnns} that adopts a complete binary (i.e., for both weights and activation) neural network (BNN) to significantly improve DNN model robustness (defined as the number of bit-flips required to degrade the accuracy to as low as a random guess). However, such an aggressive low bit-width model suffers from poor clean (i.e., no attack) inference accuracy. To counter this, we propose a novel and efficient two-stage network growing method, named \textbf{\grows}. It selectively grows the channel size of each BNN layer based on channel-wise binary masks training with Gumbel-Sigmoid function. Apart from recovering the inference accuracy, our \bnns after growing also shows significantly higher resistance to BFA. Our evaluation of the CIFAR-10 dataset shows that the proposed \bnns can improve the clean model accuracy by \textbf{$\sim$2-8 \%}, compared with a baseline BNN, while simultaneously improving the resistance to BFA by more than \textbf{125 $\times$}. Moreover, on ImageNet, with a sufficiently large (e.g., 5,000) amount of bit-flips, the baseline BNN accuracy drops to \textbf{4.3 \%} from 51.9\%, while our \bnns accuracy only drops to \textbf{37.1 \%} from \textbf{60.9 \%} (\textbf{9\%} clean accuracy improvement).
\end{abstract}


\section{Introduction}

Recently Deep Neural Networks (DNNs) have been deployed in many safety-critical applications \cite{liu2017survey}. The security of DNN models has been widely scrutinized using adversarial input examples \cite{madry2018towards,goodfellow2014explaining}, where the adversary maliciously crafts and adds input noise to fool a DNN model. Recently, the vulnerability of model parameter (e.g., weight) \cite{hong2019terminal,yao2020deephammer} perturbation has raised another dimension of security concern on the robustness of DNN model itself.

Adversarial weight attack can be defined as an attacker perturbing a target DNN model parameters stored or executing in computing hardware to achieve malicious goals. Such perturbation of model parameter is feasible nowadays due to the development of advanced hardware fault injection techniques, such as row hammer attack \cite{kim2014flipping,hong2019terminal}, laser beam attack \cite{duan2021adversarial,skorobogatov2002optical} and under-voltage attack\cite{gnad2017voltage}. Moreover, due to the development of side-channel attacks \cite{hu2020deepsniffer,batina2019csi}, it has been demonstrated that the complete DNN model information can be leaked during inference (e.g., model architecture, weights, and gradients). This allows an attacker to exploit a DNN inference machine (e.g., GPU, FPGA, mobile device) under an almost white-box threat model. Inspired by the potential threats of fault injection and side-channel attacks, several adversarial DNN model parameter attack algorithms \cite{hong2019terminal,rakin2019bit,rakin2020t,zhao2019fault,bai2021targeted} have been developed to study the model behavior under malicious weight perturbation.

\begin{figure}[t]
\centering
    \includegraphics[width=0.48\textwidth]{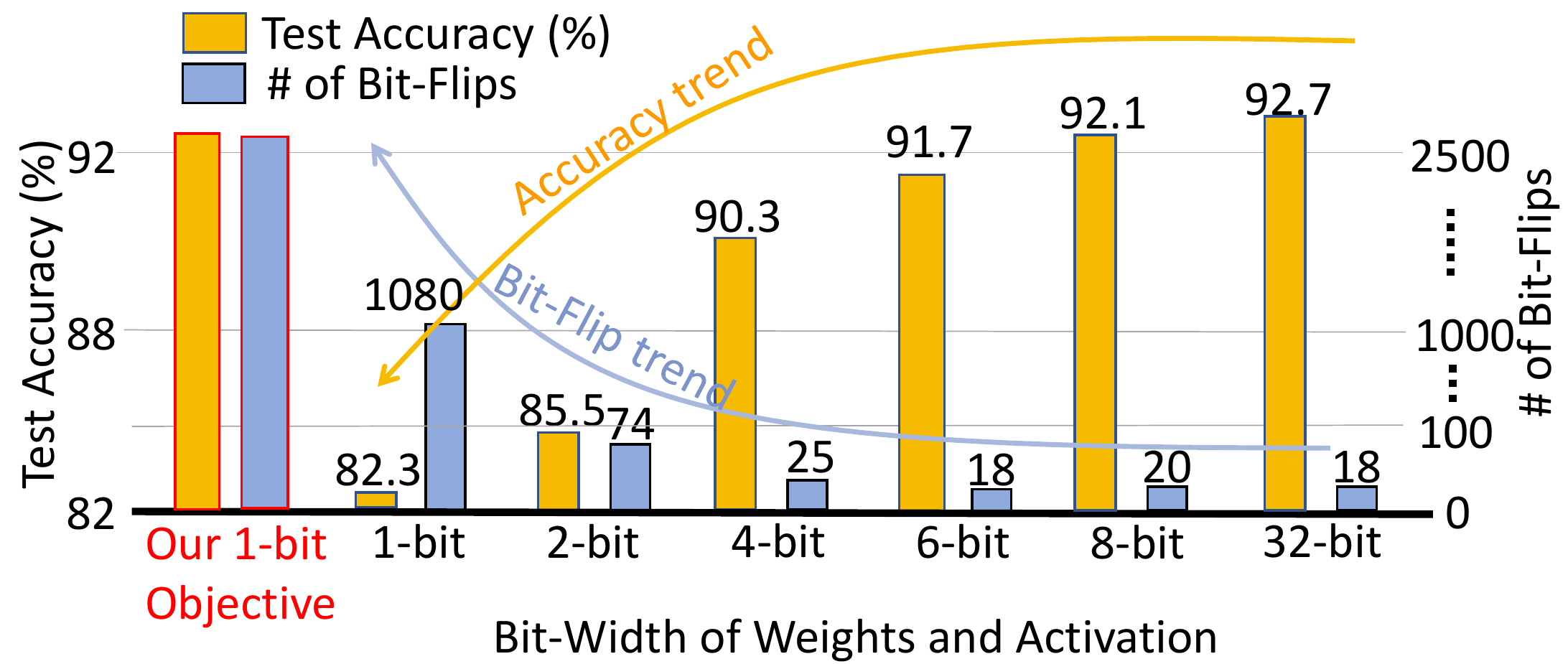} 
    \caption{\emph{The \textcolor{orange}{orange} curve shows the accuracy trend with Bit-Width of weights and activations of Resnet-20 on Cifar10. The \textcolor{blue}{blue} curve shows the trend for \# of bit-flips required to complete malfunction. On the \emph{left}, the \textcolor{red}{red} label depicts the objective of this work to simultaneously improve robustness and clean accuracy.}}
    \label{fig:trend}
  
\end{figure}

Among the popular adversarial weight attacks, bit-flip attack (BFA) \cite{rakin2019bit,yao2020deephammer,rakin2020t,bai2021targeted} is proven to be highly successful in hijacking DNN functionality (e.g., degrading accuracy as low as random guess) by flipping an extremely small amount (e.g., tens out of millions) of weight memory bits stored in computer main memory. In this work, following those prior works, we also define \textbf{\textit{robustness}} as the degree of DNN resistance against bit-flip attack, meaning a more robust network should require more number of bit-flips to hijack its function.

A series of defense works \cite{He_2020_CVPR,li2020defending,stutz2021bit} have attempted to mitigate such a potent threat. Among them, the binary weight neural network \cite{He_2020_CVPR} has proven to be the most successful one in defending BFA attacks. \cite{He_2020_CVPR} shows binarization of weight can improve model robustness by \emph{4-28 $\times$}. Note that the prior BFA defense work \cite{He_2020_CVPR} has not explored the impact of binary activation. However, for a binary weight neural network, with a sufficiently large amount of attack iterations, the attacker can still successfully degrade its accuracy to as low as random guess~\cite{He_2020_CVPR}. More importantly, due to aggressively compressing the floating-point weights (i.e., 32 bits or more) into binary (1 bit), BNN inevitably sacrifices its clean model accuracy by \emph{10-30 \%}, which is widely discussed in prior works \cite{lin2020rotated,zhou2016dorefa,liu2018bi,darabi2018regularized,courbariaux2016binarized}. 
Therefore, in this work, for the first time, our primary goal is to construct a robust and accurate binary neural network (with both binary weight and activation) to simultaneously defend bit-flip attacks and improve clean model accuracy.  


To illustrate the motivation of our work, we tested the accuracy and robustness of a ResNet-20 \cite{he2016deep} network with different bit-widths of weights and activation using CIFAR-10 dataset \cite{krizhevsky2010cifar}, following the same bit-flip attack methods in prior works \cite{rakin2019bit,yao2020deephammer,he2020defending}. 
As displayed in \cref{fig:trend}, in line with prior defense works, we observe an increase in network robustness with a lower bit-width network, especially for the case of Binary Neural Network (BNN, 1 bit) with significant improvement. Besides, BNN comes with additional computation and memory benefits which makes it a great candidate for mobile and hardware-constrained applications \cite{conti2018xnor}. As a result, many prior works \cite{courbariaux2015binaryconnect,liu2018bi,lin2020rotated,darabi2018regularized} investigated different ways of training a complete BNN. However, a general conclusion among them is that BNN suffers from heavy inference accuracy loss. In \cref{fig:trend}, we observe a similar trend where decreasing the bit-width of a given model negatively impacts the inference accuracy. 
This presents a challenging optimization problem where a lower bit-width network comes with improved robustness but at the cost of lower accuracy.  
As shown in \cref{fig:trend}, our objective in this work is to develop a binary neural network (BNN) to improve its robustness (i.e., resistance to BFA attack), while not sacrificing the clean accuracy.

To achieve this, we propose Robust \& Accurate Binary Neural Network (\emph{\bnns}), a novel defense scheme against BFA, consisting of two key components. First, we take advantage of BNN's capability in providing improved resistance against bit-flip attack, through completely binarizing both activations and weights of every DNN layer. Second, to address the clean model accuracy loss, we introduce a novel BNN network growing method, \emph{\grows}, which selectively grows the output channels of every BNN layer to recover accuracy loss. Moreover, apart from recovering accuracy, increasing the channel size can also help to resist BFA attack as discussed in \cite{He_2020_CVPR,yao2020deephammer}.



Our technical contributions could be summarized as:

\begin{itemize}
    \item \emph{First,} our proposed \emph{\bnns}, utilizing binary neural network's (BNN) intrinsic robustness improvement against BFA attack, binarizes both weight and activation of DNN model. Unlike most prior BNN related works, we binarize the weights of every layer, including the first and last layer. we refer this as \emph{complete} BNN. For activation, except the input going into first (i.e., input image) and last layer (i.e., final classification layer), all activations are binarized.
    
    \item \emph{Second,} to compensate clean model accuracy loss of a complete BNN model, we propose \emph{\grows}, a trainable mask-based growing method to gradually and selectively grow the output channels of BNN layers. Since the network growing inevitably requires large computing complexity, to improve training efficiency, our \grows follows a two-stage training mechanism with an early stop mechanism of network growth. The first stage is to jointly train binary weights and channel-wise masks through Gumbel-Sigmoid function at the early iterations of training. As the growth of binary channels converges, it goes to the second stage, where the channel growing is stopped and only the binary weights will be trained based on the network structure trained in stage-1 to further minimize accuracy loss.

    \item \emph{Finally,} we perform extensive experiments on CIFAR-10, CIFAR-100 and ImageNet datasets on popular DNN architectures (e.g., ResNets, VGG). Our evaluation of CIFAR-10 dataset shows that the proposed \bnns can improve the clean accuracy by \emph{$\sim$2-8 \%} of a complete BNN while improving the resistance to BFA by more than \emph{125 $\times$}. On ImageNet, \bnns gains \emph{9\%} clean model accuracy compared to state-of-the-art baseline BNN model while completely defending BFA (i.e., \emph{5,000} bit-flips only degrade the accuracy to around \emph{37\%}). In comparison, the baseline BNN accuracy degrades to \emph{4.3 \%} with \emph{5,000} bit-flips.   
    
    \end{itemize}



\section{Background and Related Works}
\label{sec:back}

\paragraph{\emph{Adversarial Weight Attack.}}

Adversarial input example attack \cite{madry2018towards,athalye2018obfuscated} has been the primary focus of DNN security analysis. However, the recent advancement of adversarial weight attack \cite{hong2019terminal,rakin2019bit,rakin2020tbt,yao2020deephammer} has also exposed serious vulnerability issues, where the adversary maliciously perturbs the weight parameters to compromise the performance of DNN. Among them, Bit-Flip Attack (BFA) \cite{rakin2019bit,yao2020deephammer}, is one of the most powerful attacks, which can inject malicious fault into a DNN model thorough flipping an extremely small amount of memory bits of the weight parameters stored in main memory (i.e., DRAM). For example, malicious BFA 
is demonstrated on real hardware system \cite{yao2020deephammer,hong2019terminal} through popular fault injection methodology (e.g., row-hammer \cite{kim2014flipping}). Besides, a 
recent variant of BFA 
can perform a targeted attack \cite{rakin2020t}, where the adversary can fool the DNN to predict all the inputs to a specific target class. To dive deep into BFA, 
next, we introduce the threat model of popular adversarial weight attack, attack methodology, and existing counter defense measures.

\paragraph{\emph{Threat Model.}}
Prior works \cite{hong2019terminal,rakin2019bit,yao2020deephammer} of adversarial weight attack have concluded that attacking a model with quantized weights (e.g., 8-bit) is more difficult than attacking a full-precision weight model. The underlying reason is that the weight quantized models restrict the range of weight values that the attacker can potentially manipulate. Besides, DNN quantization is proven to be extremely effective in storing and executing large DNN models in memory-constrained computing hardware. Thus, similar to the prior adversarial weight attacks \cite{rakin2019bit,yao2020deephammer}, we only focus on attacking quantized DNN model. In our attack scheme, we consider a DNN with $N_\textup{q}$-bit quantized weights. The weight matrix can be parameterized by binary quantized weight tensor $\{\tB_l\}_{l=1}^L$, where $l \in \{1, 2,...,L\}$ is the layer index. We follow a similar weight quantization and bit representation format as BFA \cite{rakin2019bit,yao2020deephammer}.

For BFA threat model, the attacker can access model parameters, architecture, and gradient information. Such a white-box threat model is becoming more and more practical due to the advancement of side-channel model information leakage attacks \cite{hu2020deepsniffer,batina2019csi}. Moreover, the attacker can easily get access to a sample test batch input ($\vx$) with correct label($\vt$). However, we assume the attacker is denied access to any training information (e.g., training data, learning rate, algorithm).

\paragraph{\emph{BFA Algorithm.}}
The Bit-Flip Attack (BFA) progressively searches for vulnerable bits by first performing a bit ranking within each layer based on gradient.
The gradient is computed w.r.t each bit of the model ($ |\nabla_{\tB_l} \mathcal{L}|$) where $\mathcal{L}$ is the inference loss function. At each iteration, the attacker performs two key attack steps: inter-layer search and intra-layer search to identify a vulnerable weight bit and flips it. Given a sample input $x$ and label $t$, the BFA \cite{rakin2019bit} algorithm tries to maximize the following loss function: 

\begin{equation}
\label{eqt:BFA}
\begin{gathered}
\max_{\{\hat{\tB}_l\}}  ~\mathcal{L}\Big (f \big( \vx ; \{\hat{\tB}_l\}_{l=1}^{L} \big), {\vt} \Big) 
\end{gathered}
\vspace{-5pt}
\end{equation}

The attacker's goal is to maximize the loss $\mathcal{L}$; while ensuring the hamming distance between the perturbed weight tensor by BFA ($\hat{\tB}_{l=1}^L$) and initial weight tensor ($\{\tB_l\}_{l=1}^L$) remains minimum.  Finally, the attack efficiency can be measured by the number of bit-flips required to cause the DNN to malfunction. Note that, BFA  \cite{yao2020deephammer,rakin2019bit} is an un-targeted attack; it can only degrade the DNN accuracy by maximizing the inference loss function. A more recent targeted variant of the attack \cite{rakin2020t} can force the DNN to classify the input samples into a specific target class by minimizing the loss w.r.t to the target class label.

\begin{equation}
\label{eqt:BFA}
\begin{gathered}
\min_{\{\hat{\tB}_l\}}  ~\mathcal{L}\Big (f \big( \vx ; \{\hat{\tB}_l\}_{l=1}^{L} \big), \hat{\vt} \Big) 
\end{gathered}
\vspace{-5pt}
\end{equation}

Here, $\hat{\vt}$ is the target class label. In this work, we consider both un-targeted and targeted BFA to evaluate \bnns.

\paragraph{\emph{Prior Defense against BFA.}}
The very first BFA known as terminal brain damage \cite{hong2019terminal} proposed weight quantization to defend against BFA. However, a recent BFA scheme \cite{rakin2019bit} demonstrated accuracy degradation of an 8-bit quantized ResNet-18 from 69 \% to 0.1 \% on ImageNet dataset with only 13 bit-flips. To counter the efficacy of BFA, weight binarization \cite{He_2020_CVPR}, piece-wise weight clustering \cite{He_2020_CVPR}, weight reconstruction \cite{li2020defending}, adversarial training \cite{stutz2021bit}, and model size increasing \cite{rakin2020t} have been investigated. Among them, binarization of the weight parameters \cite{He_2020_CVPR} has been proven to be most effective and successful in limiting the efficacy of BFA attack. However, none of the existing defense methods could completely defend BFA. In \cite{He_2020_CVPR}, it is experimentally demonstrated that binary weights can significantly improve the resistance to BFA, but the robustness gain comes at the cost of greatly reduced clean model accuracy. The primary objective of this work is to break this opposing trend (discussed in \emph{\cref{fig:trend}}) between clean model accuracy and robustness in a binary neural network.

In a tangent line of works, the possibility of defending BFA by introducing model integrity check \cite{li2021radar,9094220,zhao2021ft} and system-level protection \cite{patelblockhammer,li2020deepdyve} has also been explored. However, our goal in this work is to improve the intrinsic robustness of DNN model from an algorithm perspective without requiring system-level support. Note that, any system-level integrity check or protection scheme can be integrated on top of our algorithm-level defense to further enhance security.

\section{Defense Intuition}

The key intuition of our proposed defense \bnns is inspired by the BNN's intrinsic improved robustness against adversarial weight noise. As discussed in \cref{sec:back}, the resistance of binarized weights to BFA is a well-investigated \cite{He_2020_CVPR,rakin2020t} phenomenon. Our evaluation presented in \cref{tab:inti1} also demonstrates the similar efficacy of a binary weight model in comparison to an 8-bit quantized weight model. However, unlike prior works \cite{He_2020_CVPR}, we completely binarize DNN model including both weights and activations. Surprisingly, \cref{tab:inti1} shows that a complete BNN requires $\sim$39 $\times$ more bit-flips than an 8-bit model to achieve the same attack objective (i.e., 10 \% accuracy). 

\textbf{\emph{Observation 1.}} \textit{A complete binary neural network (i.e., both binary weights and activations) improves the robustness of DNN significantly (e.g., $\sim$39 $\times$ in \cref{tab:inti1}) }.

\begin{table}[ht]
\centering
\caption{\emph{ResNet-20 \cite{he2016deep} performance on CIFAR-10 dataset. We report the \# of bit-flips required to degrade the DNN accuracy to 10.0 \%  as the measure of robustness \cite{rakin2019bit}.}}
\label{tab:inti1}
\scalebox{0.75}{
\begin{tabular}{@{}cccc@{}}
\toprule
\begin{tabular}[c]{@{}c@{}}Model\\ Type\end{tabular} & \begin{tabular}[c]{@{}c@{}}Clean \\ Acc. (\%)\end{tabular} & \begin{tabular}[c]{@{}c@{}}Acc. After \\ Attack (\%)\end{tabular} & \begin{tabular}[c]{@{}c@{}}\# of \\ Bit-Flips\end{tabular} \\ \midrule
8-bit weight \cite{rakin2019bit} & 92.7  & 10.0 &  28 \\
Binary weight \cite{He_2020_CVPR} & 89.01 & 10.99 & 89 \\
Binary weight + activation & 82.33 & 10.0 & 1080 \\ \bottomrule
\end{tabular}}

\end{table}

\begin{figure}[ht]
    \centering
    \includegraphics[width=0.3\textwidth]{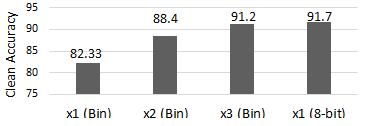} 
    \caption{\emph{Accuracy of three binary models with channel multiplier 1/2/3 and an 8-bit model (x1). Channel multiplier means output and input channel of each layer is multiplied by a constant.}}
    \label{fig:channel}

\end{figure}

As shown in \emph{\cref{tab:inti1}}, while the complete BNN may provide superior robustness, it comes at the cost of reducing the test accuracy by 10.0 \% ( from 92.7 \% to 82.33 \%) even on a small dataset with 10 classes. In \emph{\cref{fig:trend}}, we already demonstrate the trend of decreasing network bit-width can improve DNN robustness at the expense of accuracy loss. To counter this, \cite{mishra2017wrpn} has discussed a potential method of recovering the accuracy of a binary model by increasing the input and output channel width (i.e., channel multiplier) of each layer. To summarize the effect of channel multiplier in \cref{fig:channel}, we vary the BNN channel multiplier as 1/2/3 and report the clean accuracy. As the channel multiplier increases from 1 to 3, it is possible to recover the accuracy of a complete BNN model close to the 8-bit precision model. 

\textbf{\emph{Observation 2.}} \textit{By multiplying the input and output channel width with a large factor (e.g., 3), BNN accuracy degradation is largely compensated}.

Inspired by observation-1 and -2, in this work, we aim to resolve the accuracy and robustness trade-off presented in \cref{fig:trend} by constructing a DNN with complete binarization in computation level and fined-grained channel multiplication at architecture level. However, the technical challenges come from how to learn the channel multiplier (e.g., 2x, 2.5x, 3x, etc.) before training. Besides, it is also important to minimize the model size increment through optimizing layer-wise fine-grained multipliers, instead of uniform channel multiplier across whole layers. 
Therefore, in this work, our objective is to develop a general methodology to recover BNN accuracy with a fine-grained channel multiplier for each layer with little additional cost.

To achieve this, we propose a channel width optimization technique called \textit{\grows}, which is a general method of growing a given complete BNN model at early training iterations by selectively and gradually growing certain channels of each layer. 

In summary, our defense intuition is to leverage binary weights and activations to improve the robustness of DNN, while growing channel size of each individual BNN layer using our proposed \grows method to simultaneously recover the accuracy loss and further boost the robustness. \grows ensures two key advantages of our interest: i) our constructed \bnns does not suffer from heavy clean model accuracy loss, and ii) it supplements the intrinsic robustness improvement through marginally increasing BNN channel size.


\section{Proposed \bnns:}

Our proposed \bnns consists of two aspects to improve robustness and accuracy simultaneously. \textit{First}, we \textit{binarize} the weights of every DNN layer, including the first and last layers. Further, to improve the robustness, we also binarize the inputs going into each layer except the first and last one. Such complete BNN with binary weights and activation is the principal component of our defense mechanism. \textit{Second}, to recover the accuracy and further improve the robustness of a complete BNN model, inspired by the early stop strategy in Neural Architecture Search (NAS)~\cite{liang2019darts+} that stops the searching at early few training iterations to avoid overfitting and reduce searching cost, we propose \textit{\grows}, a fast and efficient way of growing a BNN with a trainable channel width of each individual layer. 

\begin{figure*}[ht]
\centering
  \includegraphics[width=0.45 \linewidth]{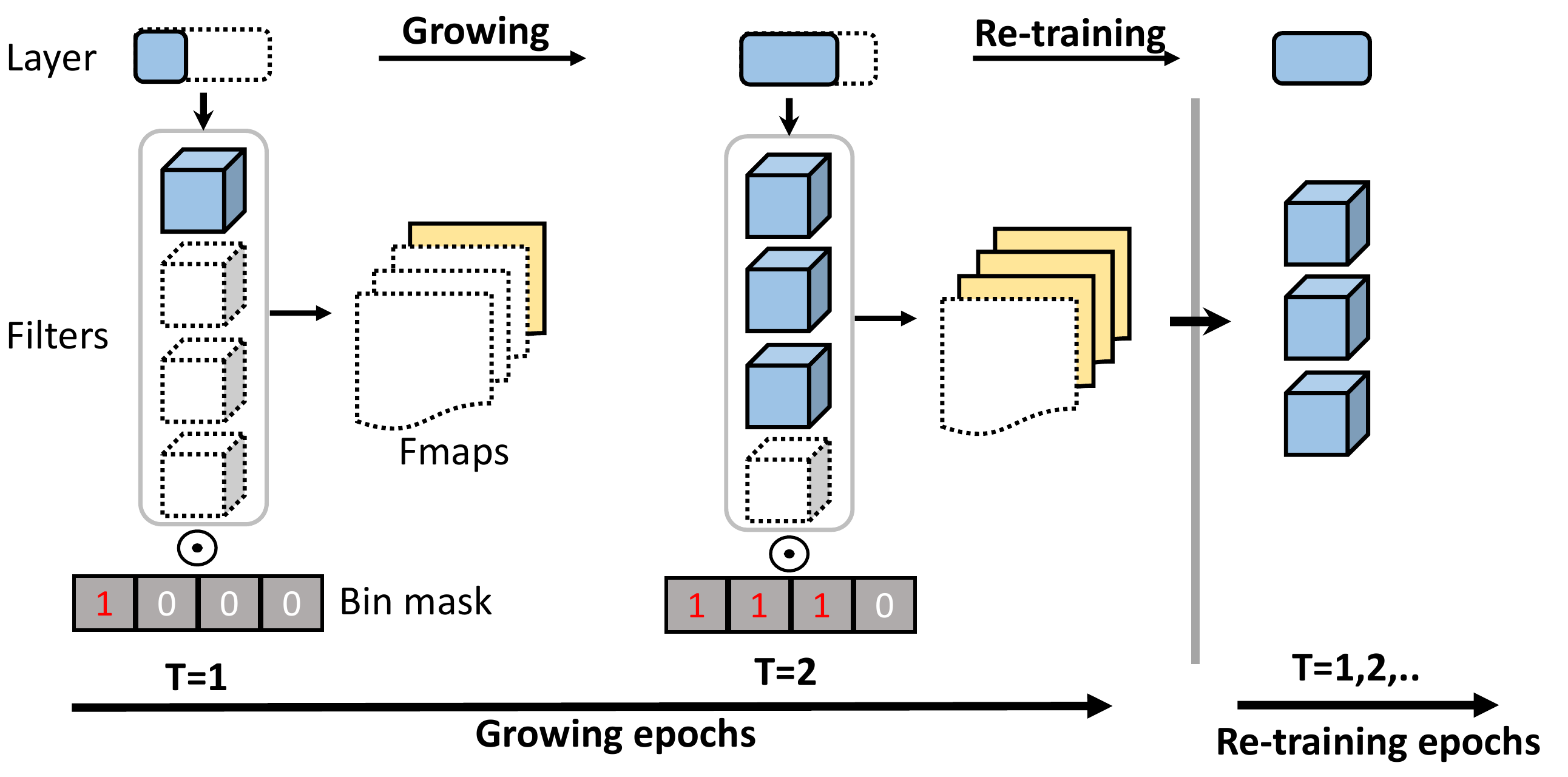}
  \includegraphics[width=0.44 \linewidth]{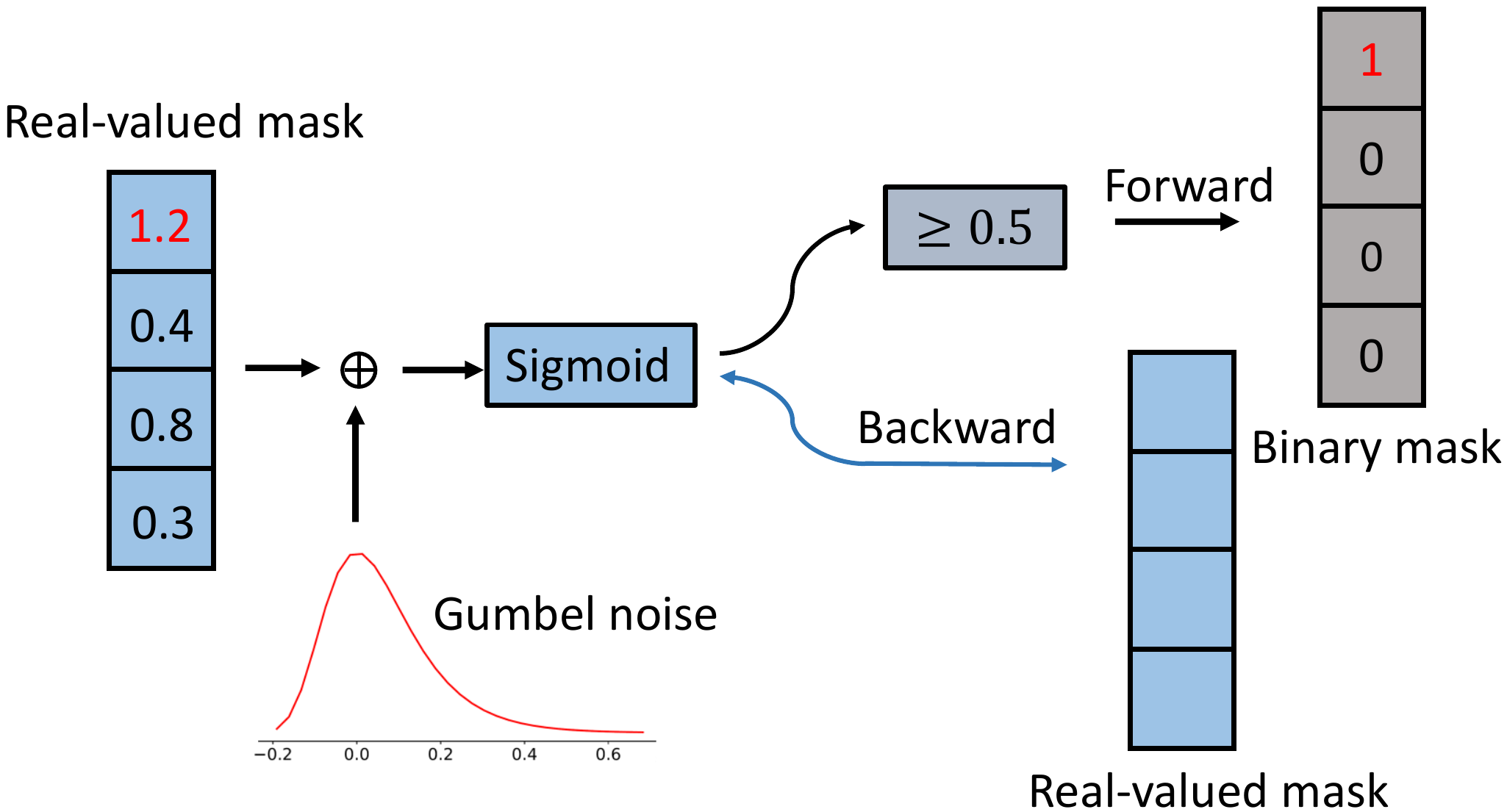}
\caption{\emph{\textbf{Left:} Detailed overview of our proposed \grows method. It is a two-stage training scheme: i) Growing: growing the output channel of every layer during the early training iterations. a new channel will be created (i.e. growing) if the associated channel mask switches from 0 to 1 for the first time. ii) Re-training: training the weight parameters based on the new BNN structure learned in stage-1. \textbf{Right:} The binary masks are trained using a combination of Gumbel-Sigmoid function and hard thresholding.}}
\label{fig:two_stage}

\end{figure*}

As shown in \cref{fig:two_stage}, \grows consists of two-stage training. In stage-1 (\textit{growing} stage), the model grows gradually and selectively from the initial baseline model to a larger model in a channel-wise manner. It is achieved by learning binary masks associated with each weight channel (growing when switching from '0' to`1' at the first time), to enable training-time growth.
As the network growth becomes stable after a few initial iterations, the growth will stop and enter stage-2. In stage-2, the new model obtained in stage-1 will be re-trained to minimize the defined loss. After completing both the growing and re-training stages, we expect to obtain a complete BNN with improved robustness without sacrificing large inference accuracy loss.

Although \cite{yuan2020growing} also utilizes the mask-based learning to grow DNN models, there are two key differences we want to highlight here: 1) \cite{yuan2020growing} jointly trains the mask and weight for the whole training process which is unstable and suffers from higher training cost. In a different way, we adopt a two-step training scheme with an early stop of growing, improving the training efficiency and thus scalability. 2) More importantly, \cite{yuan2020growing} generates the binary mask by learning full precision mask following non-differential Bernoulli sampling, which leads to the multiplication between full precision weight and full precision mask in the forward pass. This method is not efficient in our BNN training with binary weight, whereas it is much preferred to only use binary-multiplication with both operands (i.e. weight and mask) in binary format. To achieve this goal, we propose a differentialable Gumbel-Sigmoid method, as will be discussed below, to learn binary mask and guarantee binary-multiplication between weight and mask in the forward path.

\subsection{Binarization}
Our first step is to construct a complete BNN with binary weights and activations. Training a neural network with binary weights and activations \cite{courbariaux2016binarized,courbariaux2015binaryconnect,zhou2016dorefa} presents the challenge of approximating the gradient of a non-linear sign function. To compute the gradient efficiently and improve the binarization performance, in this work, we use training aware binary approximation function \cite{lin2020rotated} instead of the direct sign function:
\vspace{-10pt}
\begin{equation}
  f(\bm{z})=\begin{cases}
    k\cdot(-sign(\bm{z})\frac{t^2\bm{z}^2}{2} + \sqrt{2}\bm{z}), & \text{if $|\bm{z}|<\frac{\sqrt{2}}{t}$}.\\
    k\cdot sign(\bm{z}), & \text{otherwise}. \\
  \end{cases}
\end{equation}
\begin{equation}
  t= 10 ^ {-2 +\frac{3i}{T}}; k= max(\frac{1}{t},1) ,
\end{equation}
where $i$ is the current training iteration and T is the total iterations. At the early training stage, $\frac{i}{T}$ is low, and the above function is continuous. As the training progresses, the function gradually becomes a sign function whose gradient can be computed as:
\vspace{-8pt}
\begin{equation}
  f'(\bm{z})= \frac{\delta f(\bm{z})}{\delta \bm{z}} = max(k\cdot\sqrt{2}t - |t^2\bm{z}|,0)
\end{equation}

For weight binarization, z represents full-precision weights ($\bm{z} = \vw_{fp}$), and for activation binarization, z represents full-precision activation input going into the next convolution layer ($\bm{z} = a_{fp}$). In our \bnns, we binarize the weights of every layer including the first and last layer. However, for activation, we binarize every input going into the convolution layer except the first layer input. 

\subsection{\grows}
The goal of our proposed \grows method is to grow each layer's channel-width of a given BNN model during training to help recover the inference accuracy and further improve the robustness. As shown in \cref{fig:two_stage}, our \grows consists of two training stages: \textit{Stage-1: Growing} \& \textit{Stage-2: Re-Training}. The objective of the growing stage is to learn to grow the output channel size of each layer during the initial iterations of training. As the network architecture becomes stable, the growing stage will stop and generate a new model architecture for stage-2 to re-train the weight parameters.

\paragraph{Stage-1: Growing}
In order to gradually grow the network by increasing the output channel size, we utilize the channel-wise binary mask as indicator (i.e., on/off). Considering a convolution layer with input channel size $c_{in}$, output channel size $c_{out}$, filter size $k\times k$ and output filter $\vw^j \in \mathbf{R}^{\{ k\times k \times c_{out}\}}$; then the $j^{th}$ ($j \in \{1,2,..c_{out}\}$) output feature from a convolution operation becomes:
\vspace{-4pt}
\begin{equation}
  h_{out}^j = \textbf{Conv}(h_{in},\vw^j_{b} \odot \vm_b^j) 
  \vspace{-4pt}
\end{equation}
where $\vw_b^j \in [-1,1]$ is a binary weight ($\vw_b^j=f(\vw_{fp}^j)$), and $\vm_b^j \in [0,1]$ is a binary mask. When the binary mask is set to 0, the entire $j^{th}$ output filter is detached from the model. 
As both weight and mask are in binary format, such element-wise multiplication can be efficiently computed using a binary multiplication operation (i.e. XNOR), instead of floating-point multiplication. This is the reason why we need to guarantee both the weight and mask are in binary format in the forward path. 
The growing stage starts from a given baseline model (i.e., $\times$1 channel width) and each channel is associated with a trainable mask. 
During training, a new output filter channel will be created (i.e., growing) when the mask switches from 0 to 1 for the first time. An example of this growing procedure is illustrated in fig.~\ref{fig:two_stage}. 
The growing stage optimization objective can be mathematically formalized as: 

\begin{equation}
    \min_{\vw_{b},\vm_{b}}\mathcal{L}_{E}(g(\vw_b \odot \vm_b;\vx_{t}), \vy_{t}) 
\label{eqt:loss1}
\vspace{-4pt}
\end{equation}
where g($\cdot$) is the complete BNN inference function.
However, the discrete states (i.e. non-differential) of the binary masks $\vm_b$ make their optimization using gradient descent a challenging problem.

\vspace{-10pt}
\paragraph{\emph{Training Binary Mask.}}
The conventional way~\cite{mallya2018piggyback} of generating the binary trainable mask is to train a learnable real-valued mask ($\vm_{fp}$) followed by a hard threshold function (i.e., sign function) to binarize it. However, such a hard threshold function is not differentiable, the general solution is to approximate the gradients by skipping the threshold function during back-propagation and update the real-value masks directly. Different from that, in this work, we propose a method to eliminate the gradient estimation step and make \textbf{whole mask learning procedure differentiable}, and thus compatible with the existing gradient-based back-propagation training process. 

First, we relax the hard threshold function to a continuous logistic function:

\begin{equation}
    \sigma(\vm_{fp}) = \frac{1}{1+\textrm{exp}(-\beta\vm_{fp})},
\label{eqt:sigmoid}
\end{equation}
where $\beta$ is a constant scaling factor. 
Note that the logistic function becomes closer to the hard thresholding function for higher $\beta$ values.

Then, to learn the binary mask, we leverage the Gumbel-Sigmoid trick, inspired by Gumbel-Softmax~\cite{jang2016categorical} that performs a differential sampling to approximate a categorical random variable. Since sigmoid can be viewed as a special two-class case of softmax, we define $p(\cdot)$ using the Gumbel-Sigmoid trick as:

\begin{equation}
    p(\vm_{fp}) = \frac{ \textrm{exp}((\textrm{log} \pi_0+g_0)/T)}{\textrm{exp}((\textrm{log}\pi_0+g_0)/T) + \textrm{exp}((g_1)/T)},
\label{eqt:soft_trick}
\end{equation}
where $\pi_0$ represents $\sigma(\vm_{fp})$. $g_0$ and $g_1$ are samples from Gumbel distribution. The temperature $T$ is a hyper-parameter to adjust the range of input values, 
where choosing a larger value could avoid gradient vanishing during back-propagation. Note that the output of $p(\vm_{fp})$ becomes closer to a Bernoulli sample as $T$ is closer to 0. We can further simplify \cref{eqt:soft_trick} as:

\begin{equation}
    p(\vm_{fp}) = \frac{ 1}{1+\textrm{exp}(-(\textrm{log}\pi_0+g_0-g_1)/T)}
\label{eqt:gumb_simp}
\vspace{-2pt}
\end{equation}

Benefiting from the differentiable property of~\cref{eqt:sigmoid} and \cref{eqt:gumb_simp}, the real-value mask $\vm_{fp}$ can be embedded with existing gradient based back-propagation training without gradient approximation. During training, most values in the distribution of $p(\vm_{fp})$ will move towards either 0 or 1. To represent $p(\vm_{fp})$ as binary format, we use a hard threshold (i.e., 0.5) during forward-propagation of training, which has no influence on the real-value mask to be updated for back-propagation as shown in \cref{fig:two_stage}. 
Finally, the optimization objective in \cref{eqt:loss1} can be reformulated as:

\begin{equation}
    \min_{\vw_{fp},\vm_{fp}}\mathcal{L}_{E}(g(f(\vw_{fp}) \odot p(\vm_{fp});\vx_{t}), \vy_{t})
\label{eqt:loss}
\vspace{-3pt}
\end{equation}

\paragraph{Stage-2: Re-training}
After stage-1, we obtain the grown BNN structure (i.e., channel index for each layer), which is indicated by the channel-wise mask $\vm_b$. Then, in stage-2, we construct the new BNN model $\vw_{fp}^{*}$ using the weight channels with mask values as 1 in $\vm_b$ and discard the rest. In stage-2 training, the newly constructed BNN's weight parameters will be trained without involving any masks. 
The re-training optimization objective can be formulated as:

\begin{equation}
    \min_{\vw_{fp}^{*}}\mathcal{L}_{E}(g(f(\vw_{fp}^{*}));\vx_{t}), \vy_{t})
\label{eqt:loss2}
\vspace{-4pt}
\end{equation}

After completing the re-training stage, we expect to obtain a complete BNN with simultaneously improved robustness and clean model accuracy.

\begin{table*}[ht]
\centering
\caption{\emph{We report the Clean Accuracy (\%) and the number of weight-bits in Million (M) of ResNet-18 model on three datasets. We show \bnns accuracy improves by 3 \% on CIFAR-10, 6 \% on CIFAR-100 and 9 \% on ImageNet in comparison to baseline binary model.}}
\label{tab:accuracy}

\scalebox{0.8}{
\begin{tabular}{@{}|c|c|c|c|c|c|c|@{}}
\toprule
\multicolumn{1}{|c|}{} & \multicolumn{2}{c|}{\emph{CIFAR-10}} & \multicolumn{2}{c|}{\emph{CIFAR-100}} & \multicolumn{2}{c|}{\emph{ImageNet}} \\ \cmidrule(l){2-7} 
\multicolumn{1}{|c|}{Model Type} &
  \begin{tabular}[c]{@{}c@{}} Weight-Bits (M)\end{tabular} &
  \multicolumn{1}{c|}{\begin{tabular}[c]{@{}c@{}}Clean Accuracy (\%)\end{tabular}} &
  \begin{tabular}[c]{@{}c@{}} Weight-Bits (M)\end{tabular} &
  \multicolumn{1}{c|}{\begin{tabular}[c]{@{}c@{}}Clean Accuracy (\%) \end{tabular}} &
  \begin{tabular}[c]{@{}c@{}}Weight-Bits (M)\end{tabular} &
  \begin{tabular}[c]{@{}c@{}}Clean Accuracy (\%)\end{tabular} \\ \midrule
Full-Precision              & 357.8 (32$\times$)     & 94.2     & 362.2 (32$\times$)     & 76.2    & 374.4 (32$\times$)      & 69.7         \\
Binary (RBNN \cite{lin2020rotated}) 
& 11.18 & 90.14         & 11.32 & 66.14         & 11.7 & 51.9         \\
\textit{\bnns} \textit{(Proposed)}                  & \emph{23.37 ($\sim$ 2$\times$)}         & \emph{92.9 ( $\sim$ 3 $\uparrow$)}         &  \emph{39.53 ($\sim$ 4 $\times$)}          &         \emph{72.29 ( $\sim$ 6 $\uparrow$)}      &  \emph{73.09 ( $\sim$ 6.5 $\times$)}          &       \emph{60.9( $\sim$ 9 $\uparrow$) }       \\ \bottomrule
\end{tabular}}

\end{table*}
\begin{table*}[ht]
\centering
\caption{\emph{Summary of Robustness Analysis of CIFAR-10. The total number of Weight bits is reported in Million. The improvement factor of \bnns is listed in comparison to baseline 8-bit model.}}
\label{tab:CIFAR10}
\scalebox{0.85}{
\begin{tabular}{@{}|c|c|c|c|c|c|c|c|c|c|c|c|c|@{}}
\toprule
\begin{tabular}[c]{@{}c@{}} \emph{Model}\\ \emph{Bit}\\ \emph{Width}\end{tabular} & \begin{tabular}[c]{@{}c@{}}\emph{Total}\\ \emph{Weight}\\ \emph{Bits (M)}\end{tabular} & \begin{tabular}[c]{@{}c@{}} \emph{CA} \\ (\%)\end{tabular} & \begin{tabular}[c]{@{}c@{}}\emph{PA}\\  \emph{(\%)}\end{tabular} & \begin{tabular}[c]{@{}c@{}}\emph{\# of}\\ \emph{Bit}\\ \emph{Flips}\end{tabular} & \begin{tabular}[c]{@{}c@{}}\emph{Total}\\ \emph{Weight}\\ \emph{Bits (M)}\end{tabular} & \begin{tabular}[c]{@{}c@{}} \emph{TA}\\ \emph{(\%)}\end{tabular} & \begin{tabular}[c]{@{}c@{}}\emph{PA}\\  \emph{(\%)}\end{tabular} & \begin{tabular}[c]{@{}c@{}}\emph{\# of}\\ \emph{Bit}\\ \emph{Flips}\end{tabular} & \begin{tabular}[c]{@{}c@{}}\emph{Total}\\ \emph{Weight}\\ \emph{Bits (M)}\end{tabular} & \begin{tabular}[c]{@{}c@{}}\emph{CA}\\ \emph{(\%)}\end{tabular} & \begin{tabular}[c]{@{}c@{}}\emph{PA}\\  \emph{(\%)}\end{tabular} & \begin{tabular}[c]{@{}c@{}}\emph{\# of}\\ \emph{Bit}\\ \emph{Flips}\end{tabular} \\ \midrule
\multicolumn{13}{|c|}{\textit{Un-Targeted Attack : The classification accuracy drops down to random guess level (e.g., for a 10 class problem: 1/10 $\times$ 100 = 10\%)}} \\ \midrule
 & \multicolumn{4}{c|}{\emph{ResNet-20}} & \multicolumn{4}{c|}{\emph{ResNet-18}} & \multicolumn{4}{c|}{\emph{VGG}} \\ \midrule
\begin{tabular}[c]{@{}c@{}}8-bit\end{tabular} & 2.16 & 91.71 & 10.9 & {\color[HTML]{000000} 20} & 89.44 & 93.74 & 10.01 & 17 & 37.38 & 93.47 & 10.8 & 34 \\ \midrule
4-bit & 1.08 & 90.27 & 10.1 & {\color[HTML]{000000} 25} & 44.72 & 93.13 & 10.87 & 30 & 18.64 & 90.76 & 10.93 & 26 \\ \midrule
Binary & 0.27 & 82.33 & 10.0 & {\color[HTML]{000000} 1080} & 11.18 & 90.14 & 17.7 & 5000 & 4.66 & 89.24 & 10.99 & 2149 \\ \midrule
\emph{\bnns} & \begin{tabular}[c]{@{}c@{}} 1.94 \end{tabular} & 90.18 & 10.0 & {\color[HTML]{000000} \textbf{\begin{tabular}[c]{@{}c@{}}\emph{2519}\\ (\emph{$\times$125})\end{tabular}}} & \begin{tabular}[c]{@{}c@{}}23.37 \end{tabular} & {\color[HTML]{000000} 92.9} & {\color[HTML]{000000}  51.29} & {\color[HTML]{000000} \textbf{\begin{tabular}[c]{@{}c@{}}\emph{5000}\\ (\emph{$\times$294})\end{tabular}}} & \begin{tabular}[c]{@{}c@{}}20.6\end{tabular} & {\color[HTML]{000000} 91.58} & {\color[HTML]{000000}  72.68} & {\color[HTML]{000000} \textbf{\begin{tabular}[c]{@{}c@{}}\emph{5000}\\ (\emph{$\times$147})\end{tabular}}} \\ \midrule
\multicolumn{13}{|c|}{\textit{Targeted Attack : Classifies all the inputs to a target class. As a result, for a 10 class problem test accuracy drops to 10.0 \% }} \\ \midrule
 & \multicolumn{4}{c|}{\emph{ResNet-20}} & \multicolumn{4}{c|}{\emph{ResNet-18}} & \multicolumn{4}{c|}{\emph{VGG}} \\ \midrule
8-bit & 2.16 & 91.71 & 10.51 & 6 & 89.44 & 93.74 & 10.71 & 20 & 37.38 & 93.47 & 10.75 & 28 \\ \midrule
4-bit & 1.08 & 90.27 & 10.72 & 4 & 44.72 & 93.13 & 10.21 & 21 & 18.64 & 90.76 & 10.53 & 13 \\ \midrule
Binary & 0.27 & 82.33 & 10.99 & 529 & 11.18 & 90.14 & 10.99 & 1545 & 4.66 & 89.24 & 10.99 & 2157 \\ \midrule
\emph{\bnns} & \begin{tabular}[c]{@{}c@{}}1.94\end{tabular} & 90.18 & 10.97 & {\color[HTML]{000000} \textbf{\begin{tabular}[c]{@{}c@{}}\emph{226}\\ (\emph{$\times$37.66})\end{tabular}}} & \begin{tabular}[c]{@{}c@{}}23.37 \end{tabular} & 92.9 & 10.99 & {\color[HTML]{000000} \textbf{\begin{tabular}[c]{@{}c@{}}\emph{3983}\\ (\emph{$\times$199})\end{tabular}}} & \begin{tabular}[c]{@{}c@{}}20.6\end{tabular} & 91.58 & \emph{ 78.99} & {\color[HTML]{000000} \textbf{\begin{tabular}[c]{@{}c@{}}\emph{5000}\\ (\emph{$\times$178})\end{tabular}}} \\ \bottomrule
\end{tabular}}

\end{table*}
\section{Experimental Details}

\subsection{Dataset and Architecture}

We evaluate \bnns on three popular vision datasets: CIFAR-10 \cite{krizhevsky2010cifar}, CIFAR-100 \cite{krizhevsky2010cifar} and ImageNet~\cite{krizhevsky2012imagenet}. In our evaluation, we split the dataset following the standard practice of splitting it into training and test data \footnote{https://github.com/lmbxmu/RBNN}. We train ResNet-20 \cite{he2016deep}, ResNet-18 \cite{he2016deep} and a small VGG \cite{lin2020rotated} architecture for CIFAR-10 dataset. For both CIFAR-100 and ImageNet, we demonstrate the efficacy of our method in ResNet-18. We follow the same DNN architecture, training configuration, and hyper-parameters as rotated binary neural network (RBNN) \cite{lin2020rotated}. We also follow the same weight binarization method as RBNN including applying the rotation on the weight tensors before binarization.

\subsection{Attack and Defense Hyper-Parameters.}

Un-targeted BFA \cite{rakin2019bit} degrades the DNN accuracy close to a random guess level (1/no. of class)$\times$100 (e.g., 10 \% for CIFAR-10, 1 \% for CIFAR-100 and 0.1 \% for ImageNet). We also perform N-to-1 targeted \cite{rakin2020t} attack where the attacker classifies all the inputs into a specific target class, which again will degrade the overall test accuracy to a random guess level. We run the attack three rounds and report the best round in terms of the number of bit-flips required to achieve the desired objective. Besides, we assume the maximum \# of bits the attacker can flip is 5,000, based on practical prior experiments. \cite{gruss2018another} shows, to get around strong bit-flip  protection schemes (e.g., SGX \cite{mckeen2016intel}), requires around 95 hours to flip 91 bits using double-sided row-hammer attack. At this rate, to flip 5,000 bits, it would cost around seven months, which is practically impossible. Thus, if our defense can resist 5,000 bit flips, we can safely claim that \bnns can completely defend BFA.  We provide the hyper-parameters of the defense implementation in the appendix.
\vspace{-1em}
\subsection{Evaluation Metric}

\textbf{Clean Accuracy \% (CA)} is the percentage of test samples correctly classified by the DNN model when there is no attack. \textbf{Post-Attack Accuracy \% (PA)} is defined as the test accuracy after conducting the BFA attack on the DNN model. To measure the model robustness, we report the \textbf{Number of Bit-flips} required to degrade the DNN accuracy to a random guess level (e.g., 10.0 \% for CIFAR-10). To evaluate the DNN model size, we report the \textbf{Total Number of Weight Bits (B)} present in the model in Million.

\section{Results}

\subsection{Accuracy Evaluation}
We present the performance of our proposed \bnns method in recovering the accuracy of a BNN in \emph{\cref{tab:accuracy}}. First, for CIFAR-10, the clean accuracy drops by  \emph{3.6 \%} after binarization. However, by growing the model by \emph{2 $\times$} from a base (\emph{1 $\times$}) BNN model, our proposed \grows helps to recover the clean accuracy back to \emph{92.9 \%}. Similarly, for CIFAR-100 and ImageNet, the clean accuracy drops by \emph{$\sim$9 \%} and \emph{$\sim$18 \%} after weight and activation binarization of every layer for ResNet-18. Again, after growing the base (\emph{1 $\times$}) BNN model to a \emph{6.5 $\times$} size, \grows recovers \emph{$\sim$9 \%} of the clean accuracy on ImageNet. 

In summary, complete binarization impacts the clean accuracy of BNN by up to \emph{18 \%} degradation on a large-scale dataset (e.g., ImageNet). Thus, recovering the accuracy near the baseline full-precision model becomes extremely challenging. Moreover, unlike prior BNNs \cite{lin2020rotated,mishra2017wrpn} that do not binarize the first and last layers, in this work, we binarize every layer and achieve \emph{60.9 \%} clean accuracy. 
In the next subsection (\emph{\cref{tab:imagenet}}), we will demonstrate that our \bnns's under-attack accuracy still holds to \emph{37 \%} while all the other baseline models drop below \emph{5 \%}.


\subsection{Robustness Evaluation.}
\paragraph{\emph{CIFAR-10.}}

In \emph{\cref{tab:CIFAR10}}, we summarize the robustness analysis on CIFAR-10 dataset. First, for un-targeted attack, our proposed \bnns improves the resistance to BFA by requiring \emph{125 $\times$} more 
bit-flips for ResNet-20. As for ResNet-18 and VGG, they demonstrate even higher resistance to BFA. The attack fails to degrade the accuracy below \emph{72 \%} even after 5,000 flips on VGG architecture. Similarly, the attack fails to break our defense on ResNet-18 architecture as well. We demonstrate that even after 5,000 flips the test accuracy still holds at \emph{51.29 \%}, while it requires only 17-30 flips to completely malfunction the baseline models (e.g., 4-bit/8-bit). In conclusion, our proposed \bnns increases the model's robustness to BFA significantly, most notably completely defending BFA on the VGG model.

However, we observe that the targeted BFA is more effective against our \bnns. As \emph{\cref{tab:CIFAR10}} shows, \bnns still improves the resistance to T-BFA \cite{rakin2020t} by \emph{37 $\times$} and \emph{199 $\times$} on ResNet-20 and ResNet-18 architectures, respectively, in comparison to 8-bit model counterparts. For the VGG model, the targeted BFA can only degrade the DNN accuracy to \emph{78.99 \%} even after 5,000 bit-flips; thus completely defending the attack. In summary, BFA can degrade the DNN accuracy to near-random guess level (i.e., 10 \%) with only $\sim$\emph{4-34} bitflips. But \bnns improves the resistance to BFA significantly. Even after \emph{37-294 $\times$} more bit-flips, it still fails to achieve the desired attack goal.

\begin{table}[ht]
\centering
\caption{\emph{Robustness evaluation of CIFAR-100 and ImageNet against BFA. An 8-bit weight quantized model is used as baseline.}}
\label{tab:imagenet}
\scalebox{0.8}{
\begin{tabular}{@{}|c|ccc|ccc|@{}}
\toprule
 & \multicolumn{3}{c|}{\emph{CIFAR-100}} & \multicolumn{3}{c|}{\emph{ImageNet}} \\ \cmidrule(l){2-7} 
 & \begin{tabular}[c]{@{}c@{}}\emph{CA} \\ (\%)\end{tabular} & \begin{tabular}[c]{@{}c@{}}\emph{PA} \\ (\%)\end{tabular} & \begin{tabular}[c]{@{}c@{}}\emph{\# of} \\ \emph{Bit-}\\ \emph{Flips}\end{tabular} & \begin{tabular}[c]{@{}c@{}}\emph{CA} \\ (\%)\end{tabular} & \begin{tabular}[c]{@{}c@{}}\emph{PA} \\ (\%)\end{tabular} & \begin{tabular}[c]{@{}c@{}}\emph{\# of} \\ \emph{Bit}\\ \emph{Flips}\end{tabular} \\ \midrule
Baseline & 75.19 & 1.0 & 23 & 69.1 & 0.11 & 13 \\
Binary & 66.14 & 15.47 & 5000 & 51.9 & 4.33 & 5000  \\
\textbf{\emph{\bnns}} & \textbf{\emph{72.29}} & \textbf{\emph{54.22}} & \textbf{\emph{5000}} & \textbf{\emph{60.9}} & \textbf{\emph{37.1}} &  \textbf{\emph{5000}}\\ \bottomrule
\end{tabular}}
\end{table}

\paragraph{\emph{CIFAR-100 and ImageNet.}}
Our proposed \bnns improves the resistance to BFA on large-scale datasets (e.g., CIFAR-100 and ImageNet) as well. As presented in \emph{\cref{tab:imagenet}}, the baseline 8-bit models require less than \emph{25} bit flips to degrade the interference accuracy close to a random guess level. While binary model improves the resistance to BFA, it is still possible to significantly reduce (e.g., \emph{4.3 \%}) the inference accuracy with a sufficiently large (e.g., 5,000) amount of bit-flips. But our proposed \bnns outperforms both 8-bit and binary baseline, as even after 5000 bit flips the accuracy only degrades to 37.1 \% on ImageNet.

\begin{table}[ht]
\centering
\caption{State-of-the-Art binary ResNet-18 models on ImageNet (Top-1 \& Top-5 Clean Accuracy(\%) ).}
\label{tab:image_acc}
\scalebox{0.8}{
\begin{tabular}{|ccc|ccc|}
\toprule
\emph{Method} & \emph{Top-1 } & \emph{Top-5} & \emph{Method} & \emph{Top-1 } & \emph{Top-5} \\ \midrule
ABC-Net \cite{lin2017towards} & 42.7 & 67.6 & XNOR++ \cite{bulat2019xnor} & 57.1 & 79.9 \\
XNOR-Net \cite{rastegari2016xnor} & 51.2 & 73.2 & IR-Net \cite{qin2020forward} & 58.1 & 80 \\
BNN+ \cite{darabi2018regularized} & 53.0 & 72.6 & RBNN \cite{lin2020rotated}& 59.9 & 81.9 \\
Bi-Real \cite{liu2018bi} & 56.4 & 79.5 & \textbf{\emph{\bnns}} & \textbf{\emph{62.9}} & \textbf{\emph{84.1}} \\ \bottomrule
\end{tabular}}
\vspace{-0.75em}
\end{table}

\subsection{Comparison to Competing Method.}
We summarize our \emph{\bnns} defense performance in comparison to other SOTA BNN in \emph{\cref{tab:image_acc}} \footnote{In this table, we do not binarize first \& last layer for a fair comparison.}. We achieve an impressive \emph{62.9 \% Top-1} clean accuracy on ImageNet, beating all the prior BNN works by a fair margin. However, this improvement in clean accuracy comes at the cost of additional model size overhead (e.g., memory) which we will discuss in \cref{sec: overhead}.

\begin{table}[ht]
\centering
\caption{\emph{Comparison to other competing defense methods on CIFAR-10 dataset evaluated attacking a ResNet-20 model.}}
\label{tab:cmp}
\scalebox{0.9}{
\begin{tabular}{@{}cccc@{}}
\toprule
\begin{tabular}[c]{@{}c@{}} \emph{Models} \end{tabular} & \begin{tabular}[c]{@{}c@{}}\emph{CA}(\%)\end{tabular} & \begin{tabular}[c]{@{}c@{}}\emph{PA}(\%)\end{tabular} & \begin{tabular}[c]{@{}c@{}} \emph{Bit-Flips \#} \end{tabular} \\ \midrule
Baseline ResNet-20 \cite{rakin2019bit} & 91.71 & 10.9 & 20 \\
Piece-wise Clustering \cite{He_2020_CVPR} &  90.02& 10.09 & 42 \\
Binary weight \cite{He_2020_CVPR} & 89.01 & 10.99 & 89 \\
Model Capacity $\times$ 16 \cite{He_2020_CVPR,rakin2020t} & 93.7 & 10.0 & 49 \\
Weight Reconstruction \cite{li2020defending} & 88.79 & 10.0 & 79  \\
\textbf{\emph{\bnns}} (\textbf{\textit{proposed}}) & \textbf{\emph{90.18}} & \textbf{\emph{ 10.0}} & \textbf{\emph{2519}} \\ \bottomrule
\end{tabular}}
\vspace{-0.75em}
\end{table}

Apart from accuracy gain, our \emph{\bnns}'s major goal is to improve the robustness to BFA. Our superior defense performance is summarized in \cref{tab:cmp}, where \emph{\bnns} again outperforms all existing defenses. Even in comparison to the best existing defense binary weight model \cite{He_2020_CVPR}, BFA requires \emph{28 $\times$} more 
bit flips to break our defense.


\subsection{Defense Overhead}

\label{sec: overhead}

From the experimental results presented above (e.g., \cref{tab:CIFAR10}), it is evident that the accuracy and robustness improvement of \bnns comes at the expense of larger model size. But even after growing the binary model size, \bnns remains within \emph{26-90 \%} of the size of an 8-bit model, while achieving more than \emph{125 $\times$} improvement in robustness. To give an example, our \bnns model size increases by \emph{4 $\times$} in comparison to a baseline binary model for VGG. Still, our \bnns can achieve similar accuracy as a 4-bit model with comparable model size (\cref{tab:CIFAR10}). However, we already demonstrated in \cref{fig:trend} that a 4-bit quantized model fails to defend BFA. Similarly, our \bnns ResNet-18 model size is  2 $\times$ (same size as 2-bit) than a binary baseline model. Again, the accuracy of this model is comparable to a 4-bit model as shown in \cref{tab:CIFAR10}. \textit{Thus, in conclusion, our proposed \bnns model stays within the memory budget of a 2- to 6-bit model while improving the resistance to BFA by more than 125 $\times$ with significantly higher (2-9 \%) clean accuracy than a binary model}. 

\subsection{\grows Training Evolution}
In the appendix, we show a sample of the network growing evolution of our \grows method. A clear trend is the network growing for each layer converges at early training iterations (e.g. $\sim$35), which justifies our early stop mechanism of growing could greatly reduce training complexity and time.

\section{Conclusion}

In this work, we propose \emph{\bnns}, to defend against bit-flip attack. We utilize BNN's intrinsic robustness to bit-flips and propose \emph{\grows} method to increase the channel size of a BNN with the objective to simultaneously recover clean data accuracy and further enhance the defense efficiency. We demonstrate through extensive experiments that \bnns breaks the opposing trend between robustness and clean accuracy for low bit-width DNN models. We successfully construct a complete BNN to defend BFA with a significantly higher clean accuracy than a baseline BNN.

\bibliographystyle{unsrt}
\bibliography{egbib}

\appendix

\section{\emph{\grows training effect.}}
\begin{figure}[ht]
    \centering
    \includegraphics[width=0.35\textwidth]{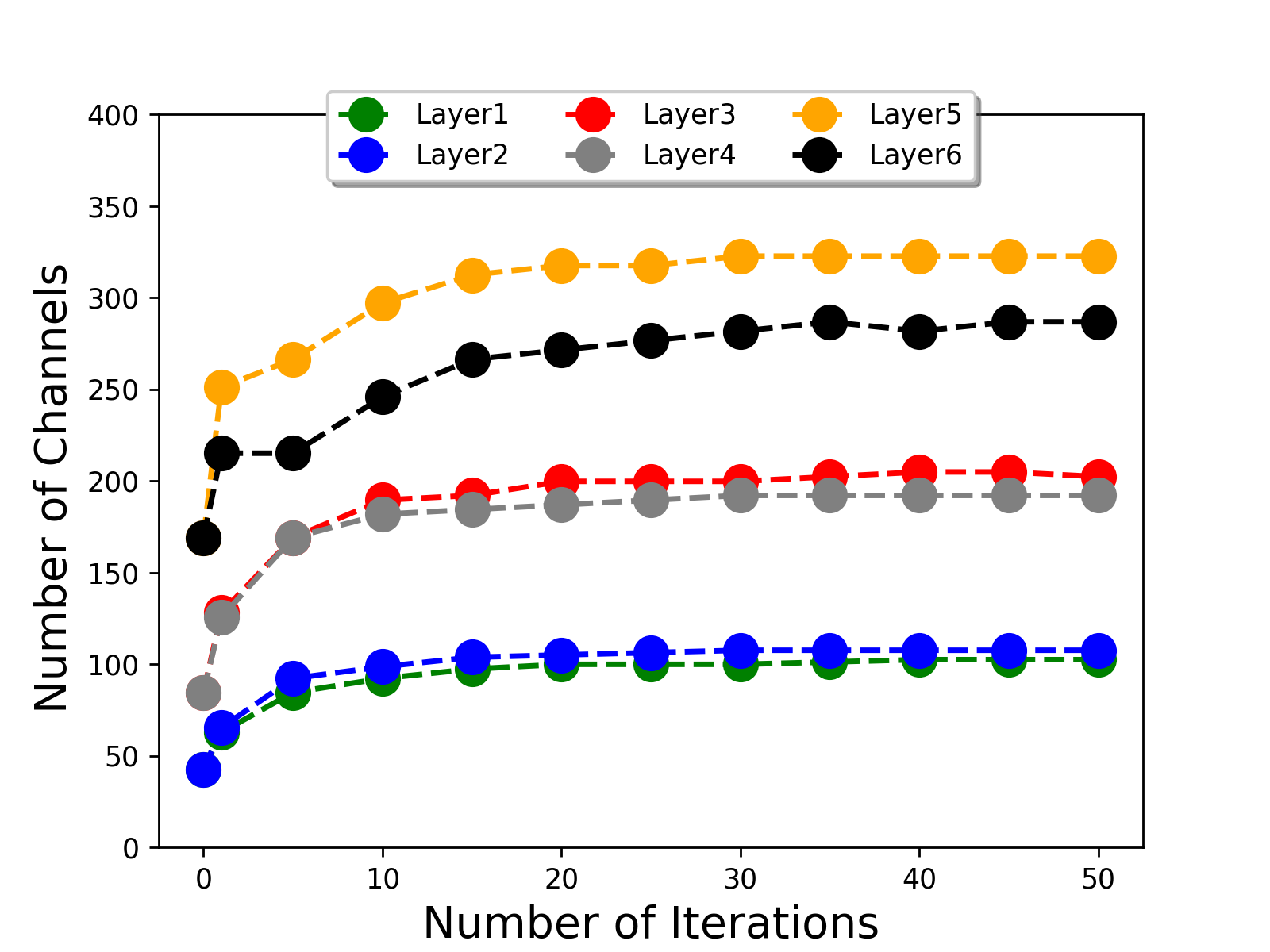} 
    \caption{\emph{The plot shows the total number of channels activated at each layer of VGG model during the first stage (growing)  training.}}
    \label{fig:channel2}
\end{figure}

In \cref{fig:channel2}, we demonstrate a sample training of \bnns where \grows increases the number of activating channels at each layer of a BNN. As the training iteration increases most of the layer channel reaches a saturation point. At around iteration no. \emph{35}, most of the layer channel size becomes stable, and the growing step is terminated. At this stage, We re-initialize the model for the re-training step. In our experiment, we terminate the growing step when most layer channel size remained fixed for two successive iterations.

\begin{figure}[ht]
    \centering
    \includegraphics[width=0.48\textwidth]{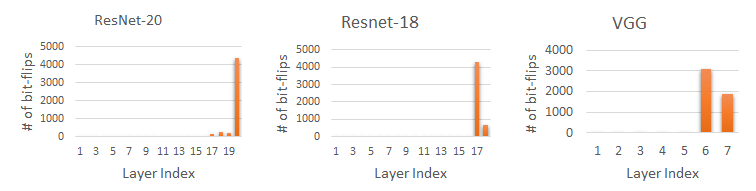} 
    \caption{\emph{Layerwise bit-flip profile of ResNet-20, VGG and ResNet-18 indicate most of the attack occurs in the last two layers.}}
    \label{fig:layerwise}
\end{figure}
\section{\emph{Layer-wise bit-flip analysis.}}
Previous BFA defense \cite{He_2020_CVPR} work have demonstrated that the majority of the bit-flips occur in the first and last layer of a DNN. However, after binarization, we observe (\cref{fig:layerwise}) that most of the bit-flips exist in the last two layers. The reason being we binarize every layer weight and activation except the input activation going into the last layer. Otherwise, all the other layer input is constrained within binary values. As a result, the input going into the last layer contains floating-point values, which is the only vulnerable point to cause significant error in the final classification layer computation by injecting faults into the last two layers. Thus binarizing the input going into the last layer can potentially nullify BFA attack even further, but training a neural network with binarized activation in the last layer is very challenging and itself presents a future research question.

\section{\emph{Defense Hyper-Parameters.}} The first, step of our training phase is to perform growing. To do so, we need to initialize the mask values; we initialize the mask values of the base model ($\times$ 1) to be equal to 1. Thus the base architecture channels always remain attached to the model at the beginning of the training. Next, we initialize the rest of the masks with a negative value within the range of -1 to -0.1 to keep their corresponding binary mask value ($m_b$) equals to 0 initially. After growing, in general, we re-initialize the model keeping only the channels with mask values equal to 1 and discarding the rest of the channels for VGG, AlexNet or other architectures w/o residual connection. However, in particular for residual models (e.g., ResNet-18) we ensure each layer within the basic-block has the same size (i.e., equal to the size of the layer with maximum channel size) to avoid output size miss-match between the residual connections and next layer.  Next, We also initialize the $\beta$ values setting them equal to 1. Besides at each iteration of the growing stage, we update the beta values using the same beta scheduler as in Eqn. 8 in  \cite{yuan2020growing}. Our defense training hyper-parameter for stage two is similar to \cite{lin2020rotated}. We will release a complete public repository of the code soon.

\end{document}

%% file: math_commands.tex
\usepackage{amsmath,amsfonts,bm}

\def\eqref#1{equation~\ref{#1}}

\def\1{\bm{1}}

\def\vm{{\bm{m}}}

\def\vt{{\bm{t}}}

\def\vw{{\bm{w}}}
\def\vx{{\bm{x}}}
\def\vy{{\bm{y}}}

\DeclareMathAlphabet{\mathsfit}{\encodingdefault}{\sfdefault}{m}{sl}
\SetMathAlphabet{\mathsfit}{bold}{\encodingdefault}{\sfdefault}{bx}{n}
\newcommand{\tens}[1]{\bm{\mathsfit{#1}}}

\def\tB{{\tens{B}}}

